# Optimal Personalised Treatment Computation through *In Silico* Clinical Trials on Patient Digital Twins[*][†]

**Stefano Sinisi** [C], **Vadim Alimguzhin, Toni Mancini, Enrico Tronci**

*Computer Science Department*
*Sapienza University of Rome, Italy*

**Federico Mari**

*Department of Movement, Human and Health Sciences*
*University of Rome Foro Italico, Italy*

**Brigitte Leeners**

*Division of Reproductive Endocrinology*
*University Hospital Zurich, Switzerland*

**Abstract.** In Silico Clinical Trials (ISCT), *i.e.*, clinical experimental campaigns carried out by means of computer simulations, hold the promise to decrease time and cost for the safety and efficacy assessment of pharmacological treatments, reduce the need for animal and human testing, and enable precision medicine.

In this paper we present methods and an algorithm that, by means of extensive computer simulation–based experimental campaigns (ISCT) guided by intelligent search, optimise a pharmacological treatment for an individual patient (*precision medicine*). We show the effectiveness of our approach on a case study involving a real pharmacological treatment, namely the *downregulation* phase of a complex clinical protocol for assisted reproduction in humans.

Address for correspondence: sinisi@di.uniroma1.it
[*]This is a greatly extended and revised version of the preliminary work presented in [1]. Authors equally contributed to this article.
[†]This article should be cited as: Stefano Sinisi, Vadim Alimguzhin, Toni Mancini, Enrico Tronci, Federico Mari, Brigitte Leeners. Optimal personalised treatment computation through in silico clinical trials on patient digital twins. Fundamenta Informaticae, 174(3-4), pages 283-310, 2020. DOI: 10.3233/FI-2020-1943
[C]Corresponding author

# Contents

## 1. Introduction

*Model-based* approaches to safety and efficacy assessment of pharmacological treatments (In Silico Clinical Trials, ISCT) hold the promise to *decrease* time and cost for the needed experimentations, *reduce* the need for animal and human testing, and enable *precision* medicine, where *personalised* treatments *optimised* for each patient can be designed before being actually administered. This is to be achieved by developing *computational models* for human physiology, patho-physiology, and drugs PharmacoKinetics (PK) and PharmacoDynamics (PD). Such computational models also define the possible physiological differences between different individuals and/or different reactions to drug administration. In particular, PK and PD models describe drug concentrations and effects over time in the human body (see, *e.g.*, [2]), respectively.

Encapsulating knowledge of the human physiology into computational models is among the goals of Virtual Physiological Human (VPH) (see, *e.g.*, [3]). Indeed, research in VPH provides mechanistic quantitative models of the human physiology at different levels of scale and for various medical areas (*e.g.*, cardiology, endocrinology, oncology and neurology, see [4, 5, 6, 7, 8]). For example, several models have been proposed for *molecules* (*e.g.*, [9]), cells (*e.g.*, [10, 11]) and single organs (*e.g.*, [12, 5, 13]). At upper levels in the hierarchy, we find models of human body *compartments*. Good examples are models in [14, 15, 16], which define the glucose regulation mechanism in patients with Type 1 Diabetes Mellitus, and the Hypothalamic–Pituitary–Gonadal (HPG) axis model presented in [6], which is specifically focused on hormones related to the human menstrual cycle. Finally, at the top level we find models of the *whole human* body, for example the Glucose-Insulin Model [17, 18], HumMod [19], and Physiomodel [20].

Independently of the level of scale used, most VPH models are defined by means of complex, highly non-linear differential equations. By means of numerical integration, such models can be executed within *simulators* in order to provide quantitative information about the time evolution of the modelled biological quantities (*e.g.*, blood hormone or glucose concentrations).

VPH models at different positions in the above hierarchy are typically defined at different levels of abstraction. In order to find a trade-off between simulation accuracy and efficiency, *multi-scale* models can be built by integrating, interconnecting, and co-simulating models of the different organs or body compartments of interest, exploiting the different available levels of abstraction as required by the focus and scope of the *in silico* analysis to be performed.

Different languages are traditionally used to define VPH models. Among them are Systems Biology Markup Language (SBML) [21] and Physiological Hierarchy Markup Language (PHML) (see, *e.g.*, [22]). The Modelica language (http://www.modelica.org), one of the most widely adopted open-standard general-purpose modelling languages for networks of dynamical systems within many areas of engineering, has been proposed as a common modelling platform for the integration of VPH, Pharmacokinetics/Pharmacodynamics (PKPD) models as well models for engineered systems such as biomedical devices (see, *e.g.*, [23, 24]) in order to facilitate their co-simulation and analysis.

### 1.1. Motivations

Current pharmacological treatments are often designed with the *average* patient in mind. A key topic in *precision medicine* is to develop pharmacological treatments *optimised* for any given individual, namely *personalised* treatments (see, *e.g.*, [25, 26]). Several optimisation criteria can be defined. A typical criterion is the minimisation of the overall amount of drug used, which often also reduces the probability of adverse side-effects and contributes to decrease the treatment cost.

With their amenability to define different individuals, VPH models of proved accuracy are a *key enabler* for precision medicine.

### 1.2. Contributions

In this paper we present methods and an algorithm that, by means of extensive computer simulation–based experimental campaigns (In Silico Clinical Trials, ISCT) guided by intelligent search, optimise a pharmacological treatment for an individual patient. We show the effectiveness of our approach on a case study involving a real pharmacological treatment, namely the *downregulation* phase of a complex clinical protocol for assisted reproduction in humans.

To do this, we exploit a state-of-the-art VPH model of the HPG axis (whose accuracy has been experimentally assessed in [6]) in order to conduct an ISCT. Namely, starting from clinical data collected from a real patient, we first compute her *digital twin*, which defines a *digital* representation of *that* patient physiology. Then, we compute *in silico*, by means of intelligent search on such a digital twin, the lightest (in terms of overall amount of administered drug) treatment still effective for that patient.

Our search algorithm extends standard artificial intelligence techniques (such as, *e.g.*, classical planning [27], CSP [28] and SAT [29]) in order to *intelligently* explore the space of possible time-series of drug administrations (treatments) by driving a simulator of the VPH model at hand. To improve the performance of our search algorithm, we also define suitable ordering heuristics (which keep our approach complete) and we evaluate their marginal performance gain. Moreover, our implementation can be easily adapted to several VPH models and to a wide class of clinical treatments as it drives a Modelica black-box simulator.

In order to assess the technical viability of our approach, we exploit the above HPG axis VPH model to conduct a multi-arm ISCT involving 21 patients (for which we have retrospective clinical measurements). For each such patient, who defines a distinct arm of our ISCT, we compute the optimal (lightest) still-effective downregulation treatment. Given that the required computation is extremely intensive, we conduct our ISCT on a large High Performance Computing (HPC) infrastructure.

The possibility to *optimise in silico* a complex treatment for a given human patient *before* its actual administration shows the potential of artificial intelligence search methods for model-based (*in silico*) medicine.

**Paper outline**

The paper is organised as follows. Section 2 describes some background knowledge, introduces the HPG axis model used in our case study and its phenotypes, defines the concept of clinical record and that of digital twin of a human patient. Section 3 describes the treatment we used as our case study. In Section 4 we describe our algorithm to find optimal personalised treatments by means of intelligent search which drives a black-box simulator of a patient digital twin. Section 5 describes our experimental campaign on retrospective data and analyses the outcome of our multi-arm ISCT. Finally, Section 6 discusses related work and Section 7 draws conclusions.

## 2. Formal framework

In this section we give the necessary background knowledge. We denote with $\mathbb{R}$, $\mathbb{R}_{0+}$, $\mathbb{R}_+$, $\mathbb{N}$, and $\mathbb{N}_+$ the sets of, respectively, all real, non-negative real, strictly positive real, non negative, and strictly positive integer numbers. Given $n \in \mathbb{N}_+$, we denote by $[n]$ the set of integers $\{1, \ldots, n\}$.

### 2.1. Virtual Physiological Human models

Virtual Physiological Human (VPH) models usually are *hybrid systems* defined by means of Ordinary Differential Equations (ODEs) (see, *e.g.*, [30, 31, 32, 33]) as stated in Definition 2.1.

**Definition 2.1.** A parametric Ordinary Differential Equation (ODE) dynamical system is defined by a set of differential equations over time (variable $t \in \mathbb{R}_{0+}$) having the following form:

$$\dot{\mathbf{x}}(t) = \mathbf{f}(\mathbf{x}(t), \mathbf{u}(t), \lambda)$$

where:

- $\mathbf{x}(t) \in \mathbb{R}^{n_x}$ is called *state vector*, and models biological quantities;

- $\mathbf{u}(t) \in \mathbb{R}^{n_u}$ is called *input vector*, and models exogenous inputs (*e.g.*, sequences of drug administrations);

- $\Lambda \subseteq \mathbb{R}^{n_\lambda}$ is called *parameter space* and $\lambda \in \Lambda$ denotes a *parameter vector*;

- $\mathbf{f}$ is a vector of functions defining system dynamics.

Above, we have $n_x, n_u, n_\lambda \in \mathbb{N}_+$.

In our setting, *administrations of $n_u$ different pharmaceutical drugs* (clinical actions) define inputs to the VPH model. Indeed, for each $i \in [n_u]$, the $i$-th component $u_i$ of the input vector $\mathbf{u}$ defines a function that associates to each time instant $t \in \mathbb{R}_{0+}$ the administered dose of the $i$-th drug, to be chosen within a set $A_i$ of possible doses (which always includes dose zero).

Given an assignment $\lambda$ to the VPH model parameters and an input vector $\mathbf{u}$ (*i.e.*, a time sequence of administrations of the $n_u$ drugs defined within the model), the associated VPH *model trajectory* is a continuous-time function $\mathbf{x}_\lambda(t)$ (or, to simplify notation, $\mathbf{x}(\lambda, t)$) representing the time evolution of the biological quantities defined by the model when fed with $\mathbf{u}$ starting from its initial state and with its parameters set to $\lambda$. Of course, as the VPH model is causal, its trajectories up to any time point $\bar{t} \in \mathbb{R}_{0+}$ only depend on the *restriction* of the input up to time point $\bar{t}$ (see, *e.g.*, [30, 34]).

## 2.2. The GynCycle Virtual Physiological Human model

In this section we describe the GynCycle VPH model, presented and evaluated in [6], which we will use as our case study.

GynCycle is an ODE-based model of the human female Hypothalamic–Pituitary–Gonadal (HPG) axis, with a special focus on the interactions and feedback mechanisms between hormones like Gonadotropin-Releasing Hormone (GnRH), Follicle-Stimulating Hormone (FSH), Luteinizing Hormone (LH), Estradiol (E2), Progesterone (P4), Inhibin A (IhA), and Inhibin B (IhB) during the different stages of the menstrual cycle.

The model defines the time evolution of overall 33 biological quantities (mostly blood concentration of hormones) and the Pharmacokinetics/Pharmacodynamics (PKPD) of several pharmaceutical drugs used in assisted reproduction (in particular, we will focus on GnRH analogues like Triptorelin) by means of highly non-linear differential equations.

For example, equations defining the time evolution of LH, and FSH are as follows:

$$\dot{\text{LH}}_{blood}(t) = \frac{\text{LH}_{\text{pit}}(t)}{V_{blood}} \cdot \left(b_{Rel}^{\text{LH}} + k_{\text{G-R}}^{\text{LH}} \cdot H^+(\text{G-R}(t), T_{\text{G-R}}^{\text{LH}}; n_{\text{G-R}}^{\text{LH}})\right)$$
$$- (k_{on}^{\text{LH}} \cdot R_{\text{LH}}(t) + k_{cl}^{\text{LH}}) \cdot \text{LH}_{blood}(t)$$

$$\dot{\text{FSH}}_{blood}(t) = \frac{\text{FSH}_{\text{pit}}(t)}{V_{blood}} \cdot \left(b_{Rel}^{\text{FSH}} + k_{\text{G-R}}^{\text{FSH}} \cdot H^+(\text{G-R}(t), T_{\text{G-R}}^{\text{FSH}}; n_{\text{G-R}}^{\text{FSH}})\right)$$
$$- (k_{on}^{\text{FSH}} \cdot R_{\text{FSH}}(t) + k_{cl}^{\text{FSH}}) \cdot \text{FSH}_{blood}(t)$$

In the differential equations above, LH($t$), FSH($t$) etc. denote biological *species* of interest (in the example, the blood concentration of LH and FSH, respectively), while $b_{Rel}^{\text{LH}}$, $k_{cl}^{\text{LH}}$, etc. denote *model parameters*.

As such, GynCycle is a *hybrid system* whose inputs represent drug administrations. As it happens with VPH models of practical interest as GynCycle, the number and complexity of their differential equations makes symbolic analysis (by, *e.g.*, a model checker for hybrid systems) an unviable option. Indeed, the model can be *simulated* by means of numerical integration in order to compute the time evolution of the hormones of interest upon administration of a sequence of drug doses. Thus, GynCycle is used as an *executable* black-box model.

## 2.3. GynCycle Virtual Phenotypes

VPH models like GynCycle typically take into account inter-subject variability (*i.e.*, the physiological differences among different individuals) by including suitable *parameters* in their equations (like $b_{Rel}^{\text{LH}}$ and $k_{cl}^{\text{LH}}$ mentioned above). Different value assignments to model parameters yield different model time evolutions and/or different reactions to drug administrations, thus defining different Virtual Phenotypes (VPs). Intuitively, each VP represents a class of indistinguishable (as long as the VPH model is concerned) patients. Computing a complete set of VPs for the model at hand is the starting point to obtain a *representative* population of virtual patients (hence, ideally showing all possible phenotypes). In turn, this is the key enabler to perform In Silico Clinical Trials (ISCT) to assess, *e.g.*, safety and/or efficacy of a pharmacological treatment.

Unfortunately, computing the complete set of VPs defined by a complex VPH model like GynCycle is all but easy from the computational point of view, and may require weeks of massive simulation on large High Performance Computing (HPC) infrastructures. This is due to the hidden presence of inter-dependency constraints among elements of the parameter vector which are usually not explicitly known, and hence not modelled. By ignoring such constraints (which would happen by, *e.g.*, choosing values for the parameter vector elements in a uninformed random way) would yield with high probability to model behaviours that violate the laws of biology (and hence such parameter assignments could not be regarded as VPs).

In [35, 36], a statistical model checking–based approach using intelligent global search heuristics has been discussed. Such a methodology enables the computation of a population of VPs (parameter assignments showing time evolutions compatible with the laws of biology) that is complete with respect to the used heuristics up to an arbitrarily high statistical confidence. This approach was recently generalised in [37], where it proved its effectiveness in computing a population of VPs for a VPH model of the glucose regulation system in patients with Type 1 Diabetes Mellitus.

In this paper, we conduct our ISCT starting from a population $\mathcal{P}$ of around $10^7$ VPs, extracted using the algorithm in [35, 36] from a huge search space of $10^{75}$ possible candidate parameter vector assignments (set $\Lambda$ in Definition 2.1). By exploiting *model locality* assumptions typically made on ODE-based VPH models (*i.e.*, small changes in the model parameters are expected to yield small changes in the overall model trajectory), the domain of each of the 75 dimensions of $\Lambda$ has been discretised into 10 possible values (see [35] for an evaluation of the appropriateness of this choice). We then ran the parallel tool described in [36] with error threshold equal to $0.01\%$ and statistical confidence bound equal to $95\%$. This means that, upon termination, we have $95\%$ confidence that the probability of finding new VPs by further searching is at most $0.01\%$.

### 2.4. Clinical records

In order to compute personalised optimal treatments, we will start from clinical data collected from a human patient. Along the lines of [35], a clinical record $\mathcal{C}$ (Definition 2.2) consists of a sequence of measurements for all biological quantities of interest, where each measurement maps a time instant to a measured value.

**Definition 2.2. (Clinical record)**
Let $S_1, \ldots, S_p$ be the biological quantities of interest. A patient clinical record, $\mathcal{C}$, is a tuple $(C_1, \ldots, C_p)$, where each $C_i$ ($i \in [p]$) is a finite sequence of pairs $(t, v)$ (possibly empty) where $t \in \mathbb{R}_{0+}$ denotes a time instant and $v \in \mathbb{R}_{0+}$ is the measured value of biological quantity $S_i$.

In our case study (fertility treatments), clinical records of interest collect measurements for the main hormones involved in the human menstrual cycle, namely: LH, FSH, E2, and P4.

## 2.5. Digital twin of a human patient

To compute a personalised treatment for a human patient, we first need to compute a *digital representation* for her. This will be done by using clinical data (in the form of a clinical record $\mathcal{C}$, Definition 2.2) available from that patient in order to select, from our representative population of VPs, the subset of VPs that are *compatible* with (*i.e.*, fit) such data.

To do so, we proceed as follows. Let $\mathcal{C}$ be a clinical record and $\lambda$ be a VP of our population. We define $\eta(\mathcal{C}, \lambda)$ as the average (over all biological quantities $S_i$, $i \in [p]$) Mean Absolute Percentage Error (MAPE) between the clinical measurements of $S_i$ in $\mathcal{C}$ and time evolutions of $S_i$ in VP $\lambda$. Formally:

$$\eta(\mathcal{C}, \lambda) \;=\; \frac{1}{p} \sum_{i=1}^{p} \frac{1}{|C_i|} \sum_{(t,v_i) \in C_i} \left| \frac{v_i - x_i(\lambda, t)}{\zeta(v_i)} \right|$$

where $\zeta(v_i)$ is $v_i$ if $v_i \neq 0$ and a small positive constant otherwise (to avoid division by zero).

By using such a metric, Definition 2.3 defines the concept of *patient digital twin* as the set of *all* VPs $\lambda$ in our population $\mathcal{P}$ having a value for $\eta(\mathcal{C}, \lambda)$ up to a given threshold.

**Definition 2.3.** Given a clinical record $\mathcal{C}$, our complete population of VPs $\mathcal{P}$, a threshold $\delta \in \mathbb{R}_{0+}$ for our error metric $\eta$, the *digital twin* $P$ of $\mathcal{C}$ is the following set:

$$P(\mathcal{C}) = \{\lambda \in \mathcal{P} \mid \eta(\mathcal{C}, \lambda) \leq \delta\}.$$

Intuitively, the digital twin of a human patient is a *digital* representation of the patient physiology in the form of *all* VPs entailing model behaviours that fit the patient clinical measurements (up to an average MAPE $\delta$).

## 3. Case study: downregulation in assisted reproduction treatments

In this section we define the treatment we used as our case study: the downregulation phase of an assisted reproduction protocol currently administered at the Department of Reproductive Endocrinology of University Hospital Zurich.

At each menstrual cycle of a human female, among the several follicles initially present in the ovaries, only one, unless in exceptional cases, grows and reaches maturity. A complex competition among follicles takes place, driven by a hormone-based feedback loop, in order to inhibit the maturation of multiple follicles.

An assisted reproduction treatment aims at neutralising (through drugs) such a feedback loop (*downregulation phase*), in order to achieve a controlled and an as-simultaneous-as-possible growth of 5–15 ovarian follicles (*stimulation phase*). When the first three follicles reach a size where a mature oocyte (*i.e.*, an egg cell ready for fertilisation) can be expected (about 18 mm in diameter), ovulation is induced. Then, the oocytes are retrieved, those which are mature are tried to be fertilised *in vitro*, and implanted either immediately within the treatment cycle or later after cryopreservation (*i.e.*, a preservation process by cooling) back into the uterus.

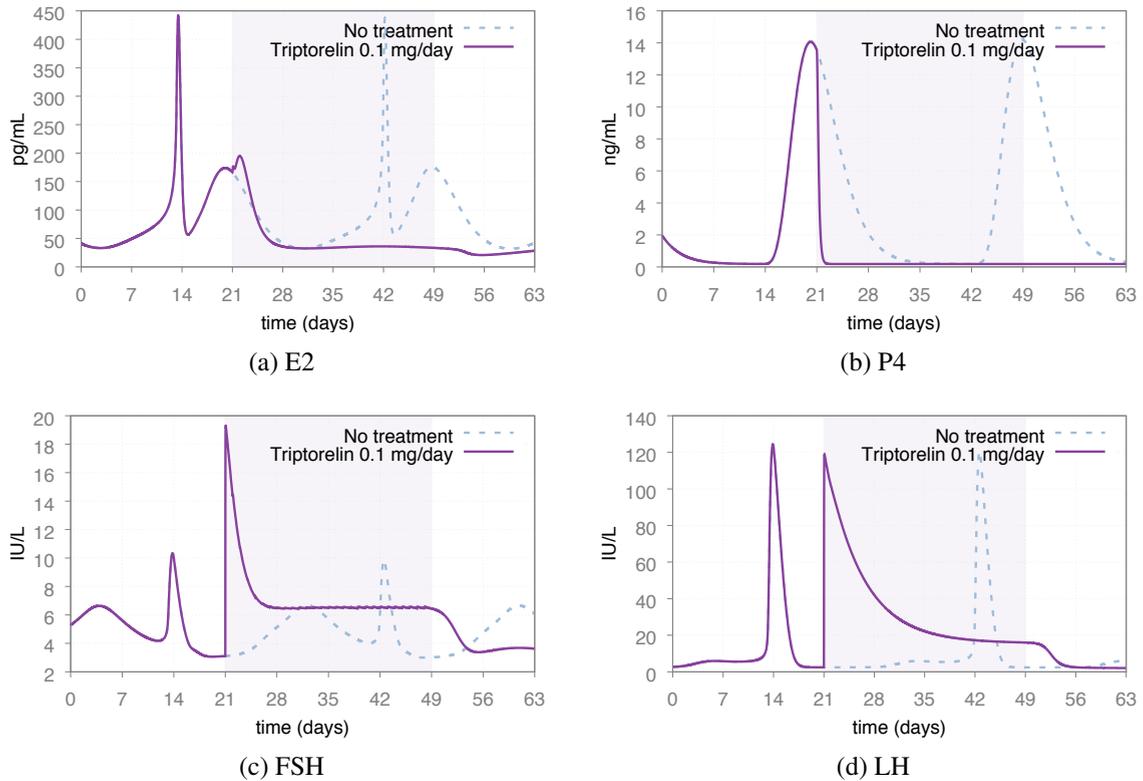

Figure 1: Expected downregulation effects on some of the hormones playing a role in the human female menstrual cycle. The coloured area defines the time window during which Triptorelin is administered.

Assisted reproduction treatments are complex and challenging, with low average success rates (around 30%) even in the top clinics, and with many factors that, to date, can be hardly kept under full control. Indeed, as hormonal regulatory systems occur within a complex network of endocrinological, neurological and psychological factors [38, 39, 40], they are difficult to capture within clinical studies, and model-based approaches might be of great aid in taking these many factors under better control.

Our case study is one of the worldwide classically used downregulation protocols aiming at suppressing the usual hormonal oscillations of the menstrual cycle (as in Figure 1) and preparing the patient to the following stimulation.

At the Department of Reproductive Endocrinology of University Hospital Zurich, this protocol currently consists of a sequence of daily administrations of 0.1 mg of Triptorelin, a Gonadotropin-Releasing Hormone (GnRH) analogue. For physiological reasons, downregulation is started within a precise time window of the menstrual cycle, namely between day 21 and day 25. The treatment might have different duration in order to address different patient reactions. In particular, a downregulation treatment is considered *successful* (effective) for a patient in case the blood concentrations of a given set of hormones and other physiological quantities go below certain thresholds within 9 days from the first drug administration, and stay always below such thresholds for the following 21 days. As a consequence, a downregulation treatment might last up to 30 days.

## 4. Computing optimal personalised treatments

In this section, we define the conditions of a successful treatment through invariants and goals (Section 4.1), describe how to perform digital twin simulations (Section 4.2), and describe the algorithm used at the core of our optimal personalised treatment computation (Section 4.3).

### 4.1. Modelling treatment invariants and goals

We define conditions that must be *always* and *eventually* satisfied by a successful treatment by means of *invariants* and *goals*, respectively. Invariant and goal treatment conditions are modelled as continuous-time *monitors* embedded within the Virtual Physiological Human (VPH) model, along the lines of [41, 42]. Monitors observe the state of the system and check whether the properties of interest are satisfied.

In particular, given a model trajectory $\mathbf{x}(\lambda, t)$ under a given input $\mathbf{u}$, our monitor output is UNDET as long as $\mathbf{x}(\lambda, t)$ satisfies the invariants (in other words, the monitor decision is "undetermined" as long as there is *hope* to extend the current treatment into a successful treatment) and goes to and *stays* at value FAIL as soon as invariants are *violated*. When a UNDET model trajectory satisfies the goal conditions, the monitor output turns to SUCCESS, informing the caller that the input $\mathbf{u}(t)$ defines a *successful* treatment (*i.e.*, an *effective* treatment which *always* satisfies invariants).

The use of continuous-time monitors embedded in the VPH model gives us a flexible way to model both bounded safety and bounded liveness properties, and is a standard approach in the simulation-based verification of cyber-physical systems (see, *e.g.*, [43, 44]).

In our case study, the properties of interest are the conditions of successful downregulation treatments (Section 3). In particular, our *invariant* requires that the day of the first drug administration is between day 21 and day 25 of the menstrual cycle. Moreover, the value of all the biological quantities under observation must go and stay below their thresholds from the 9-th day after the first drug administration. Our *goal condition* instead requires that values for those quantities stay below their thresholds for 21 consecutive days.

As a consequence, our monitor output is UNDET in the initial state, and eventually turns either to FAIL or to SUCCESS (see Figure 2). Model output turns to value FAIL as soon as, from the 9-th day after the first drug administration, the value of *any* of the physiological quantities under observation exceeds its threshold. Conversely, model output turns from value UNDET to value SUCCESS as soon as all the thresholds are satisfied for 21 consecutive days.

### 4.2. Digital Twin simulation

As anticipated in Section 2, the complexity of the (highly non-linear) differential equations typically occurring in actual VPH models hinders the possibility for their symbolic analysis. Numerical *simula-*

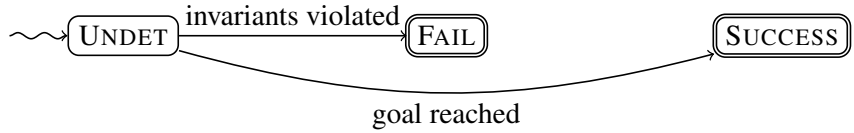

Figure 2: Treatment monitor.

*tion* is often the only means to compute the time evolution of the model quantities of interests (mainly blood hormone concentrations in our case study GynCycle model) upon a sequence of clinical actions (administration of one or more drugs with their associated doses). Also, as VPH model equations are often stiff, simulation can be an *expensive* process from a computational point of view.

Our algorithm for optimal personalised treatment computation regards the input VPH model as an *executable* black-box model. In order to drive the VPH model simulator, our algorithm assumes that it accepts the following basic commands (which are available or can be readily implemented within most modern simulators [45, 46]): `store`, `load`, `free`, `run`, `get`. Command `store`($l$) stores in memory the current state of the simulator and labels with $l$ such a state. Command `load`($l$) loads into the simulator the stored state labelled with $l$. Command `free`($l$) removes from the memory the state labelled with $l$. Command `run`($a, t$) (with $a \in A_1 \times \cdots \times A_{n_u}$ and $t \in \mathbb{R}_+$) executes clinical action $a$ (thus administering given –possibly zero– doses for all available drugs) and then advances the simulation of time $t$. Command `get`($x$) (with $x$ denoting one of the model variables) returns the value of model variable $x$ in the current state.

Since the input of our algorithm is a digital twin $P(\mathcal{C})$ computed from a patient clinical record $\mathcal{C}$ and consisting of a set of $n_p \in \mathbb{N}_+$ Virtual Phenotypes (VPs) (see Section 2.5), during the search we drive the simulation of $n_p$ *independent copies* of our VPH model (each one connected to a copy of the monitor checking for treatment invariants and goals, as explained in Section 4.1). The different model copies are instantiated with different values for the model parameter vector defined in $P(\mathcal{C})$, hence represent the different VPs in the patient digital twin.

By means of commands `run`($a, t$) ($a \in A_1 \times \cdots \times A_{n_u}$), we simultaneously apply the same clinical action to *all* VPs in $P(\mathcal{C})$ and synchronously advance all models by time $t$. Furthermore, by means of `store` and `load` commands, we can efficiently expand a tree of sequences of clinical actions, as typically done by backtracking-based state exploration algorithms. In particular, upon a backtrack, we only need to load back a previously stored state of our models, thus avoiding to simulate the entire input sequence starting from the initial state.

### 4.3. Intelligent backtracking–driven simulation

In this section we describe our intelligent search that drives a digital twin simulator, by defining our search tree (Section 4.3.1) and criteria for the evaluation and expansion of search tree nodes (Section 4.3.2). We also introduce two ordering heuristics aiming at efficiently pruning the search tree (Section 4.3.4) and speeding up the digital twin simulation time (Section 4.3.5), while retaining completeness of the overall algorithm.

### 4.3.1. Search tree definition

Given a patient clinical record $\mathcal{C}$, its digital twin $P(\mathcal{C})$, and possible alternative doses for each treatment drug at hand (*i.e.*, sets $A_1, \ldots, A_{n_u}$, each one including dose zero), our backtracking-based algorithm performs a search in the space of possible treatments (*i.e.*, time sequences of clinical actions in $A_1 \times \cdots \times A_{n_u}$). Under the following assumptions: (i) the sets of allowed doses $A_1, \ldots, A_{n_u}$ for the $n_u$ drugs are *finite*; (ii) drug administrations occur at time instants multiple of a given *time-quantum* $\tau \in \mathbb{R}_+$; and (iii) sought treatments have a bounded duration ($\tau h$, with $h \in \mathbb{N}_+$ being the treatment *horizon*), we can limit our search to a *finite tree* of maximum depth $h$, where each tree node at depth $0 \leq l \leq h$ defines a *discrete sequence* of $l$ clinical actions (elements of $A_1, \ldots, A_{n_u}$), each one occurring in the first $l$ time points multiple of $\tau$.

As described in Section 4.2, our search algorithm drives a simulator containing multiple copies of the VPH model (one per VP in the digital twin given as input), each one connected to a copy of the monitor which checks for treatment invariants and goals.

Accordingly, each node of our search tree keeps also track of the digital twin simulator state reached by injecting the clinical action sequence of that node. This is done via a `store` command (see Section 4.2). A transition between a parent node and a child node consists of extending the clinical action sequence of the parent node with a new action $a \in A_1 \times \cdots \times A_{n_u}$. In doing so, we also advance the digital twin simulator state of the parent node via a `run` command (see Section 4.2). In Section 4.3.2, we describe more in details how a search tree node is evaluated and expanded.

Before the search starts, for each VP $\lambda$ in the input digital twin, we prepare its associated copy of the VPH+monitor model within the digital twin simulator, by setting the parameter vector to $\lambda$. We also store the initial simulator state via a `store` command (see Section 4.2).

Clearly, the assumptions above together with the entailed search tree structure hold for a very wide class of treatments. For example, in our downregulation treatment case study, there is only one possible drug, namely Triptorelin, hence $n_u = 1$. Also, as such treatments allow at most one Triptorelin administration per day, we can safely set $\tau = 1$ day. Finally, successful treatments cannot last more than $h = 25 + 9 + 21 = 55$ days, starting from day 1 of the patient menstrual cycle (*i.e.*, the latest cycle day, 25, when the treatment might start, plus the latest day, 9, within which the safety conditions must be met, plus the number of consecutive days, 21, in which the safety conditions must be always satisfied in order to declare success).

Moreover, in our case study, the initial simulator state represents day 1 of the menstrual cycle of all VPs in the input digital twin.

### 4.3.2. Search tree node evaluation and expansion

At each node of the search tree, our algorithm needs to extend the current sequence of clinical actions (a treatment *prefix*, which is the empty sequence in the initial state). To do so, the algorithm:

1. Stores the current states of the VP models by means of a `store` command.

2. For each clinical action $a \in A_1 \times \ldots \times A_{n_u}$ (according to the heuristic ordering discussed in Section 4.3.4), injects $a$ into the simulator and advances the simulation (simultaneously for all VPs defined within the simulator) by time $\tau$ (this is done by issuing a command `run(a, τ)`).

3. If the monitor output for one VP returns FAIL, the algorithm infers that there is no hope to extend the partial treatment to a complete treatment successful for all the VPs in the digital twin and issues a backtrack, by loading back (via a `load` command) the simulator state of all VPs associated to the parent node in the search tree.

4. Otherwise, if the monitor output for all VPs is SUCCESS, the algorithm infers that the current treatment is successful for all VPs.

5. Otherwise, the algorithm continues by expanding the new search tree node and extending the current treatment prefix.

Whenever no further actions can be executed in the current node of the search tree, the current simulator state is freed from memory by a `free` command before triggering a backtrack.

In order to reduce the required simulation time when expanding each node, the multiple model copies are not actually packaged into a single (much heavier to be simulated) model, but are kept separate and simulated *sequentially*. This allows our algorithm to perform two performance optimisations: a sort of *early pruning* of the current treatment prefix and a *dynamic revision of the order* with which the VPs in the input digital twin will be actually simulated in the sequel (Section 4.3.5).

### 4.3.3. Optimality constraint

Since we seek the *optimal* personalised treatment for the input digital twin, our algorithm does not stop at the first found successful treatment. Indeed, along the lines of [47], it keeps track of the *lightest* successful treatment found so far, *i.e.*, the one envisioning the administration of the *minimum* overall drug amount $D_{\min} \in \mathbb{R}_{0+}$. Initially, $D_{\min}$ is set to $+\infty$.

At each node of the search tree, any action extending the current partial treatment with an administration of a drug dose which would make the overall administered drug amount reaching or exceeding $D_{\min}$ is regarded as not applicable.

The optimality constraint (whose threshold is updated each time a new optimal treatment is found), the presence of a bounded horizon $h$ and the fact that our algorithm does not stop at the first successful treatment clearly guarantee that a global optimum is always returned (if any successful treatment exists).

### 4.3.4. Action ordering heuristic

In a black-box setting as ours, no inference can be made on the effects of each candidate action without performing a simulator run to actually advance the model and then querying the model monitor output. Hence, approaches to compute, through inference, a dynamic preference order among the candidate actions to be tried during search (as those exploited in, *e.g.*, classical planning, planning for white-box hybrid systems –see Section 6– or (Q)CSP, SAT or local search solvers, see *e.g.*, [28, 29, 48, 49, 50, 51, 52]) cannot be applied.

The only information available to the algorithm without running the simulator is value $D_{\min}$ and the overall cost of the *current* partial treatment (overall amount of drug administered). Hence, given that our algorithm searches for a successful treatment of minimum cost and that any action has a non-negative contribution to the overall treatment cost, not surprisingly tries the clinical actions on each node in the search tree in ascending order of their associated dose (*i.e.*, cost), hence performing a *greedy*, optimistic choice.

### 4.3.5. Dynamic VP simulation ordering

When, during the evaluation of search tree node, the monitor output after the simulation of a VP in the digital twin returns FAIL, our algorithm, by performing a sort of *early pruning*, does not simulate nor evaluate the remaining VPs, but it backtracks and tries the next available action from the parent node, if any. Upon backtracking, the algorithm changes the verification ordering of the VPs in the digital twin in such a way that those VPs not satisfying treatment invariants or not yet simulated in the previous search step are simulated *first*. The rationale is that those VPs that already satisfied treatment invariants in previous step, will more likely satisfy treatment invariants in the current step, where a slightly different drug dose is chosen.

## 5. Multi-arm In Silico Clinical Trial

In this section we outline how we set up and conducted our In Silico Clinical Trial (ISCT) in order to assess the technical viability of the proposed approach.

### 5.1. Retrospective clinical records

Our ISCT has been conducted against retrospective clinical data courtesy of Pfizer Inc. and Hannover Medical School (Germany).

In particular, such our overall dataset consists of 21 patient clinical records for a total of around 800 measurements. Each clinical record comprises measurements for the hormones that play the most important role during the human menstrual cycle: LH, FSH, E2, and P4. Figure 3 shows measurements of such hormones in our overall dataset.

Each of the 21 clinical records in our retrospective clinical data defines a *distinct arm* of our multi-arm ISCT.

### 5.2. Computing digital twins: exclusion criteria and VP selection

From the GynCycle model, we generated a population $\mathcal{P}$ of around $10^7$ Virtual Phenotypes (VPs) as explained in [35, 36] and recapped in Section 2.3. Hence, such a population is representative of (almost) all patient behaviours allowed by the model (in the sense of [35, 36]). Note that, although generating such a population required weeks of High Performance Computing (HPC) computation, this is a one-time task for each Virtual Physiological Human (VPH) model.

In order to carry out our multi-arm ISCT, we computed a digital twin for each of the 21 clinical records in our retrospective clinical data. In particular, we set the threshold value $\delta$ (Definition 2.3) to 35% and, for each clinical record $\mathcal{C}$, we included in the digital twin of $\mathcal{C}$, $P(\mathcal{C})$, all $\lambda \in \mathcal{P}$ such that $\eta(\mathcal{C}, \lambda) \leq 35\%$. Hence, for each clinical record, we selected *all* VPs having an average Mean Absolute Percentage Error (MAPE) at most $35\%$.

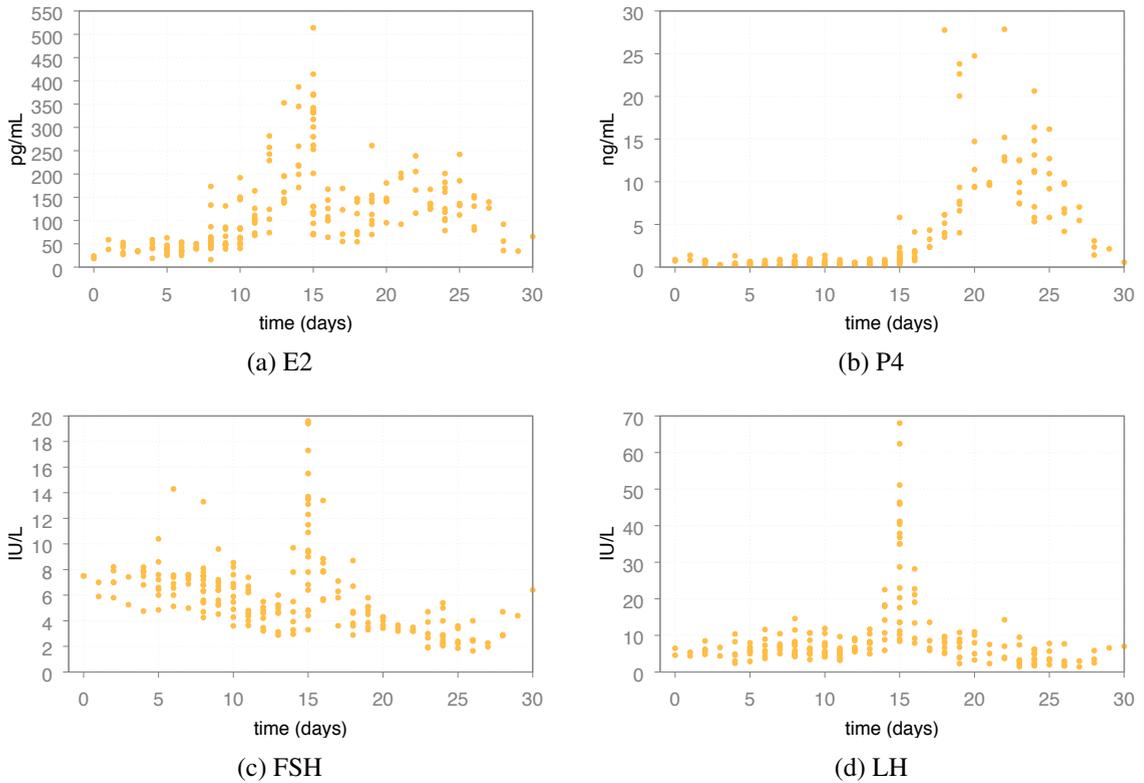

Figure 3: Measurements (dots) of, respectively, Estradiol (E2), Progesterone (P4), Follicle-Stimulating Hormone (FSH) and Luteinizing Hormone (LH) of the 21 clinical records within the dataset from Pfizer Inc. and Hannover Medical School (Germany).

Since the reference downregulation treatment of Section 3 does not succeed on all possible phenotypes (this is not surprising since, as described in Section 3, fertility treatments are known to have a low rate of success), we removed from $P(\mathcal{C})$ those VPs for which the treatment using the maximum allowed dose (*i.e.*, $0.1$ mg on each day for up to $30$ days, see Section 3) fails, for any $\mathcal{C}$. Such an *exclusion criterion* directly stems from domain knowledge on downregulation treatments (namely: for these protocols, if a treatment envisioning one full drug dose per day fails for a patient, a lighter treatment will fail as well). Our 21 digital twins contain, on average, $843\,424$ VPs each.

Unfortunately, in the worst-case scenario, each step of our backtracking search requires to simulate all VPs within the input digital twin in order to check the monitor output (*i.e.*, treatment invariants and goals, see Section 4.3.2). Thus, since simulating one VP requires seconds of numerical integration, the total computation time of our algorithm is highly affected by the size of the input digital twin, and may easily become impractical.

To overcome such an obstacle, we note that within a digital twin of a given clinical record we can (and typically do) have VPs exhibiting very similar model behaviours (both with and without drug administrations). Simulating such very similar VPs would then simply be a waste of computation as long as our ISCT is concerned. In order to reduce the size of a digital twin, we compute a subset $\hat{P}(\mathcal{C})$ of $P(\mathcal{C})$, for each given clinical record $\mathcal{C}$, such that $\hat{P}(\mathcal{C})$ does not contain such *redundant* VPs.

To do so, we define the *distance* among the time evolutions of any biological quantity $S_i$ ($i \in [n_x]$) of two VPs normalised against the time evolution of $S_i$ of a third (reference) VP (Definition 5.1).

**Definition 5.1.** Let $\lambda$, $\lambda'$ and $\lambda^\star$ be three VPs of a VPH model defining $n_x$ biological quantities.

For every biological quantity $S_i$ ($i \in [n_x]$) the *distance* between $\lambda$ and $\lambda'$ with respect to $\lambda^\star$ on $S_i$ is:

$$d_{i,\lambda^\star}(\lambda, \lambda') = \frac{\sqrt{\int_0^{+\infty} [x_i(\lambda, t) - x_i(\lambda', t)]^2 \, \mathrm{d}t}}{\zeta\left(\sqrt{\int_0^{+\infty} x_i(\lambda^\star, t)^2 \, \mathrm{d}t}\right)}$$

where, as usual, $\zeta(\alpha)$ is $\alpha$ when $\alpha \neq 0$ and a small constant otherwise (in order to avoid division by 0).

Intuitively, Definition 5.1 defines the *Euclidean distance* between the time evolutions of biological quantity $S_i$ of two given VPs ($\lambda$ and $\lambda'$) *normalised* with respect to the $L_2$-norm of that of a *reference* VP, $\lambda^\star$. In practice, we will compute the above distance using a bounded-horizon simulation, *i.e.*, model trajectories will be defined for $0 \leq t \leq h$, where $h$ is the treatment horizon. Hence, to match the above general definition, we assume that, for any $t > h$, $x_i(\lambda^\star, t)$, $x_i(\lambda, t)$ and $x_i(\lambda', t)$ are 0.

Given this notion of distance, we build $\hat{P}(\mathcal{C})$ from the digital twin $P(\mathcal{C})$ of a clinical record $\mathcal{C}$ as follows:

- We select, from $P(\mathcal{C})$, the reference VP $\lambda^\star$ as one that *minimises* $\eta(\mathcal{C}, \lambda^\star)$. In other words, $\lambda^\star$ is one of the VPs of $\mathcal{P}$ that *best matches* the clinical measurements of $\mathcal{C}$.

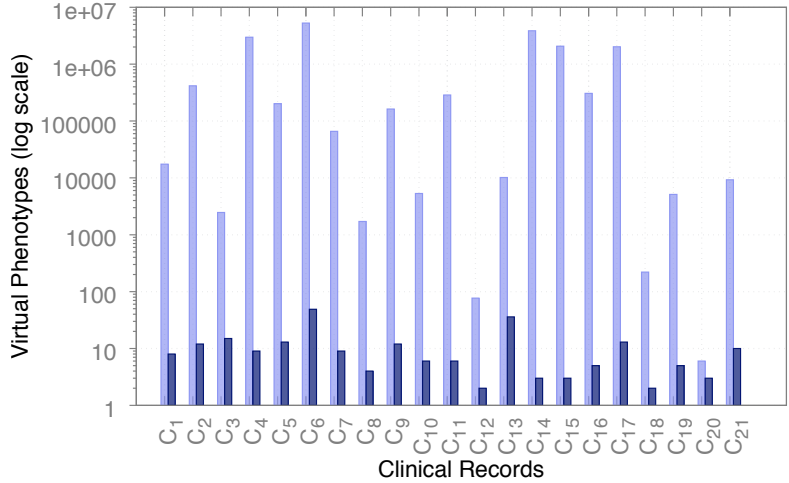

Figure 4: Bar chart showing digital twin sizes (in terms of number of VPs) before (light) and after (dark) the removal of redundant VPs and the application of our exclusion criteria for each clinical record.

- We *remove* from $P(\mathcal{C})$ those VPs whose time evolutions of *all* biological quantities have a distance below a certain threshold from other VPs in $\hat{P}(\mathcal{C})$ with respect to $\lambda^\star$ (Definition 5.2).

Such a distance threshold obviously depends on the VPH model and treatment subject of the ISCT.

**Definition 5.2.** Let $\mathcal{C}$ be a patient clinical record, $P(\mathcal{C})$ be its digital twin, $\lambda^\star$ be any VP such that $\eta(\mathcal{C}, \lambda^\star)$ is minimal, and $d$ be the distance function of Definition 5.1.

Then, given a threshold $\theta$, we call *compact digital twin* $\hat{P}(\mathcal{C})$ any subset of $P(\mathcal{C})$ such that the following condition holds:

$$\forall \lambda \in P(\mathcal{C}) \,.\, \lambda \notin \hat{P}(\mathcal{C}) \to \exists \lambda' \in \hat{P}(\mathcal{C}) \,.\, \forall i \in [n_x] \; d_{i,\lambda^\star}(\lambda, \lambda') \leq \theta/2.$$

As a result of the application of the above approach and of our exclusion criteria, we obtained compact digital twins with an average size of 11. This has been achieved by tuning the distance threshold $\theta$ in such a way that: (i) VPs having redundant and undistinguishable behaviours have been filtered out; (ii) the computed compact digital twins were representative enough for all VPs in corresponding patient digital twins; and (iii) their sizes were small enough to be computationally affordable. Figure 4 shows the distribution of their sizes.

### 5.3. Multi-arm In Silico Clinical Trial run

We ran our 21-arm ISCT using a large HPC infrastructure (the Marconi cluster) kindly provided by the Cineca consortium.

For each patient clinical record $\mathcal{C}$ in our retrospective clinical data (see Section 2.4), we, first, computed a compact digital twin $\hat{P}(\mathcal{C})$ (see Section 5.2), and then we ran our algorithm of Section 4 on an independent node of the cluster searching for an optimal (*i.e.*, lightest) and robust (with respect to *all* VPs of $\hat{P}(\mathcal{C})$) downregulation treatment for *that* specific digital twin. Thus, the whole 21-arm ISCT has been conducted in an *embarrassing parallel fashion*.

Each VP in a compact digital twin has been encoded in a Modelica (http://www.modelica.org) model (encompassing the GynCycle VPH model taking clinical actions as input, a parameter vector assignment, and a monitor to check for treatment invariants and goals) and has been simulated using JModelica v2.1 (http://www.jmodelica.org). Our search algorithm (which drives a Modelica simulator per VP) has been implemented in Python.

In the following sections, we first evaluate the marginal impact of our heuristic ordering of actions (Section 5.3.1) and of our dynamic simulation ordering of VPs within a digital twin (Section 5.3.2). Then, we present computational results (Section 5.3.3) and outcomes (Section 5.3.4) of our multi-arm ISCT.

### 5.3.1. Evaluation of the action ordering heuristic

The goal of this section is to assess the marginal impact of the heuristic presented in Section 4.3.4, which sorts actions by their ascending associated cost (administered drug dose), in the overall efficiency of the optimal personalised treatment search algorithm.

To this end, we compare it against 10 runs of an action selection strategy that evaluates the possible actions in *random* order when expanding each node of the search tree, and compare the performance of our heuristics against such reference strategy.

In particular, we defined a limit of $10^4$ on the number of search tree nodes to expand, and analysed how fast the cost (overall amount of drug dose) associated to the personalised treatments found *decreases* during the search space exploration, on all digital twins involved in our multi-arm ISCT.

In Figure 5, we present 3 representative executions where our heuristic finds an optimal solution at different points in time (with respect to the number of expanded search tree nodes). For convenience of presentation, the initial value of $D_{min}$ in Figure 5 is set to 3.0 (drug quantity employed by the reference treatment, *i.e.*, 0.1 mg on each day for up to 30 days) instead of $+\infty$ (as stated in Section 4.3.3). This is in agreement with our exclusion criteria of Section 5.2 as the reference treatment succeed on all VPs within our digital twins.

We note that, by choosing actions randomly, the algorithm might show its first solution improvements a bit earlier (Figures 5a and 5c). However, in all cases, our heuristics finds the optimal solution (*i.e.*, the lightest treatment) within a number of expanded nodes which is *orders of magnitude* lower than the number of expanded nodes required by the random action ordering strategy. As an example, Figure 5c shows that, after $10^4$ expanded nodes, 8 out of 10 runs with the random action ordering still failed to find a solution better than the reference treatment (*i.e.*, 3.0 mg).

### 5.3.2. Evaluation of the dynamic ordering of VPs within a digital twin

In order to evaluate the marginal impact of our dynamic VP simulation order within the input digital twin (Section 4.3.5) at each search tree node, we compared the execution of our algorithm against a reference obtained by averaging 10 different runs where the initial order of the VPs is randomised and fixed at the beginning of the search.

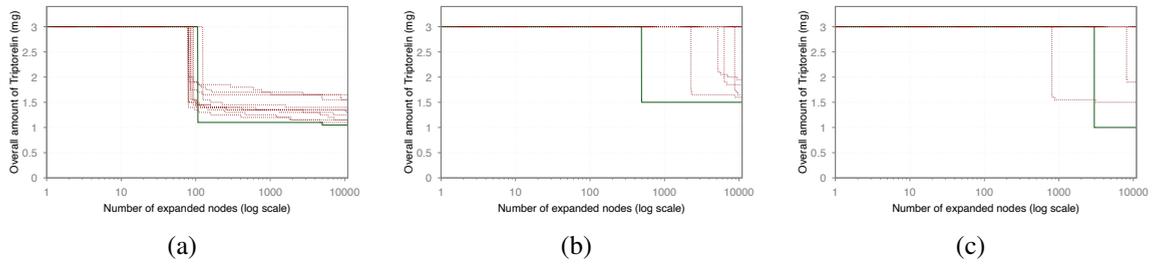

Figure 5: Comparison of solution paths traversed by our algorithm when using our ordering heuristic (green line) against the references (red dashed lines) among different digital twins of our multi-arm ISCT.

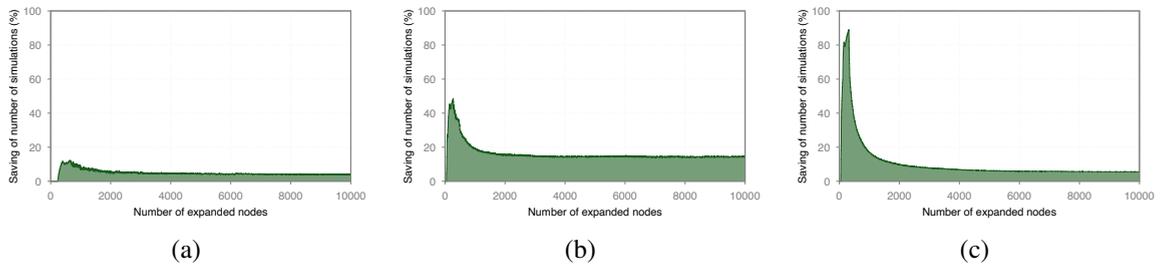

Figure 6: Marginal impact of our dynamic VP simulation ordering strategy among the digital twins of our multi-arm ISCT.

In particular, we ran our algorithm for each digital twin involved in our multi-arm ISCT, and compared the saving in terms of the number of simulations performed within each given number of search tree nodes.

Figure 6 shows our result on 3 digital twins having different sizes: (i) a small size — 3 VPs (Figure 6a); (ii) an average size — 13 VPs (Figure 6b); and (iii) a large size — 36 VPs (Figure 6c); which are representative of the spectrum of behaviours among our digital twins.

We note that the our dynamic VP simulation order is always beneficial. The highest saving in the number of simulations always occurs at the early stages of the search, where the most pruning activity is performed. In deeper stages of the search, the savings stabilise at values of the order of 10%–20%. Since the time needed to simulate a VP is essentially constant (as VPs differ only in model parameter values and not in the system of Ordinary Differential Equations, ODEs), such savings translate in comparable reductions of the overall computation time.

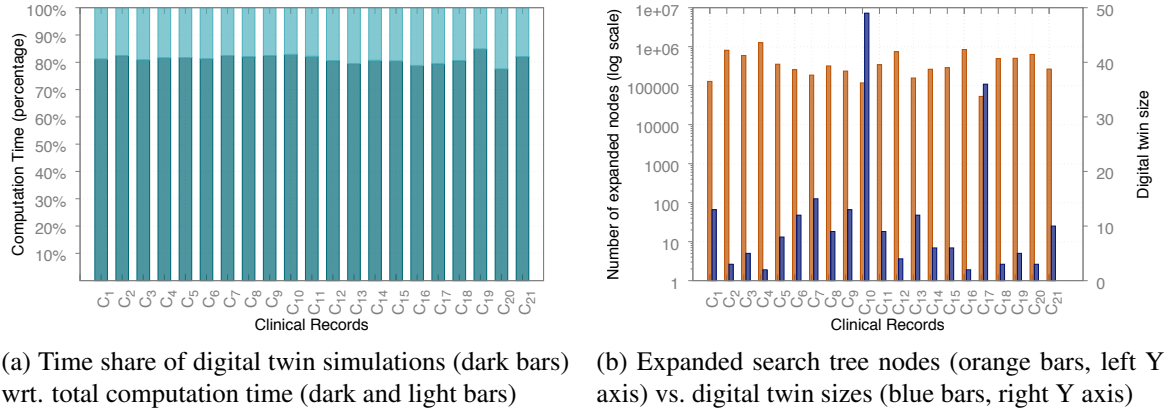

(a) Time share of digital twin simulations (dark bars) wrt. total computation time (dark and light bars)

(b) Expanded search tree nodes (orange bars, left Y axis) vs. digital twin sizes (blue bars, right Y axis)

Figure 7: Computational results of our ISCT.

### 5.3.3. Computational results

We ran our multi-arm ISCT on the Marconi HPC infrastructure kindly provided by Cineca, Italy. Due to CPU-hours budget limit, we ran each arm of our ISCT in parallel for 15 days. Each arm searches the optimal treatment for one of the 21 patients in our dataset. Indeed, our algorithm can be regarded as an *any-time* algorithm in that, at any moment during search, it keeps track of the lightest successful treatment found so far.

Figure 7a gives, for each ISCT arm, the share of simulation time within the whole computation time. As we already argued in Section 5.2, the size of the input digital twin highly impacts the speed (in terms of expanded search tree nodes per seconds) of our algorithm. In fact, in each search node (in the worst case), our algorithm needs to numerically simulate all VPs within the input digital twin. As shown in Figure 7a, the time spent in carrying out simulations is, on average, the 81.20% of the total computation time.

This implies that any technique aimed at reducing the average number of simulations per search node (like the reduction of the size of the considered digital twin defined in Section 5.2, the action order heuristic evaluated in Section 5.3.1, and the dynamic VP simulation ordering evaluated in Section 5.3.2) is expected to have a *substantial impact* on the actual performance of the algorithm.

Figure 7b shows, for each of the 21 parallel search processes (one per clinical record), the number of search nodes expanded within our time limit of 15 days (orange bars, left Y axis), and compares it to the size of the digital twin considered for that clinical record (blue bars, right Y axis).

The figure show a *clear negative correlation* (correlation coefficient equal to $-60\%$) between number of expanded nodes within our time limit and the digital twin size for all clinical records. For example, we note that when the digital twin size is small, as for, *e.g.*, $\mathcal{C}_4$, the number of expanded nodes is large, *i.e.*, near $10^6$. Instead, when the digital twin size is large, as for, *e.g.*, $\mathcal{C}_{17}$, the number of expanded nodes is quite low, *i.e.*, less than $3 \times 10^4$.

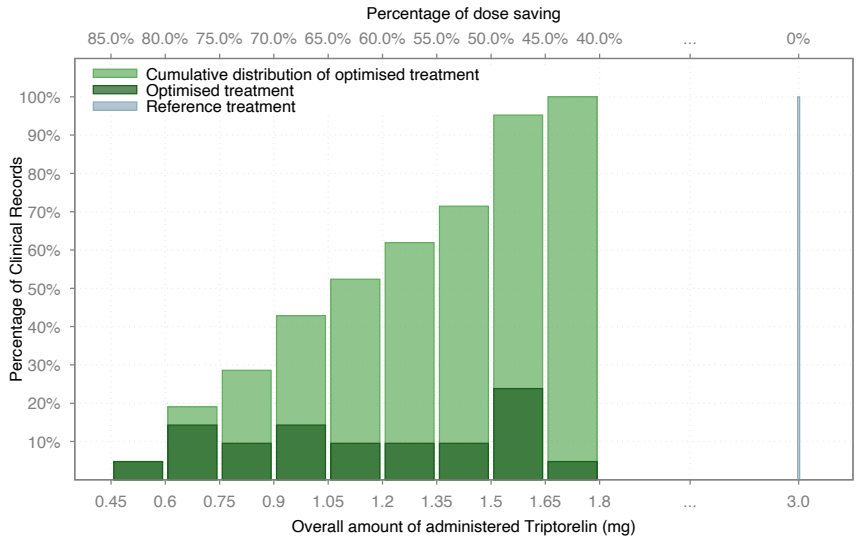

Figure 8: Multi-arm ISCT outcomes.

### 5.3.4. ISCT outcomes

Figure 8 shows the distribution of the total amount of drug employed by the optimal personalised treatments computed by our algorithm. Note that, although we ran our multi-arm ISCT with a limited time budget and did not find provably *lightest* treatments, the treatments we actually computed use, on average, less than *half* ($40.82\%$) of the drug quantity employed by the reference treatment of Section 3 (*i.e.*, $0.1$ mg on each day for up to 30 days, totalling up to 3 mg and never less than 2.8 mg, since the required thresholds are never reached before day 7). As Figure 8 shows, the computed treatments save, on average, $59.18\%$ of the drug administered in the reference treatment (standard deviation: $13\%$).

In Figure 9, we show our computed treatments for the 21 patient clinical records during our multi-arm ISCT. We note that, the majority of the computed drug administration sequences is close to the reference downregulation treatment in terms of administration frequency and treatment duration, *i.e.*, around 28 days. Besides this, there are also few computed treatments having a very light overall amount of drug dose. This of course depends on the ability of the digital twins to capture patient peculiarities and in general on the ability of the VPH model at hand to represent the human physiology of interest.

## 6. Related work

Individualised treatments have the potential value to reduce costs and improve outcomes of standard clinical treatments. In recent years data-driven techniques have been investigated thanks to the availability of big data [53]. For example, the knowledge-base approach in [54] has been used to optimise treatment plans for lung cancer. Unfortunately, in presence of scarce clinical data for the patient at hand, the above approaches cannot be applied. For example, in our case study hormones blood con-centrations are not measured every day, since those measurements are costly and invasive.

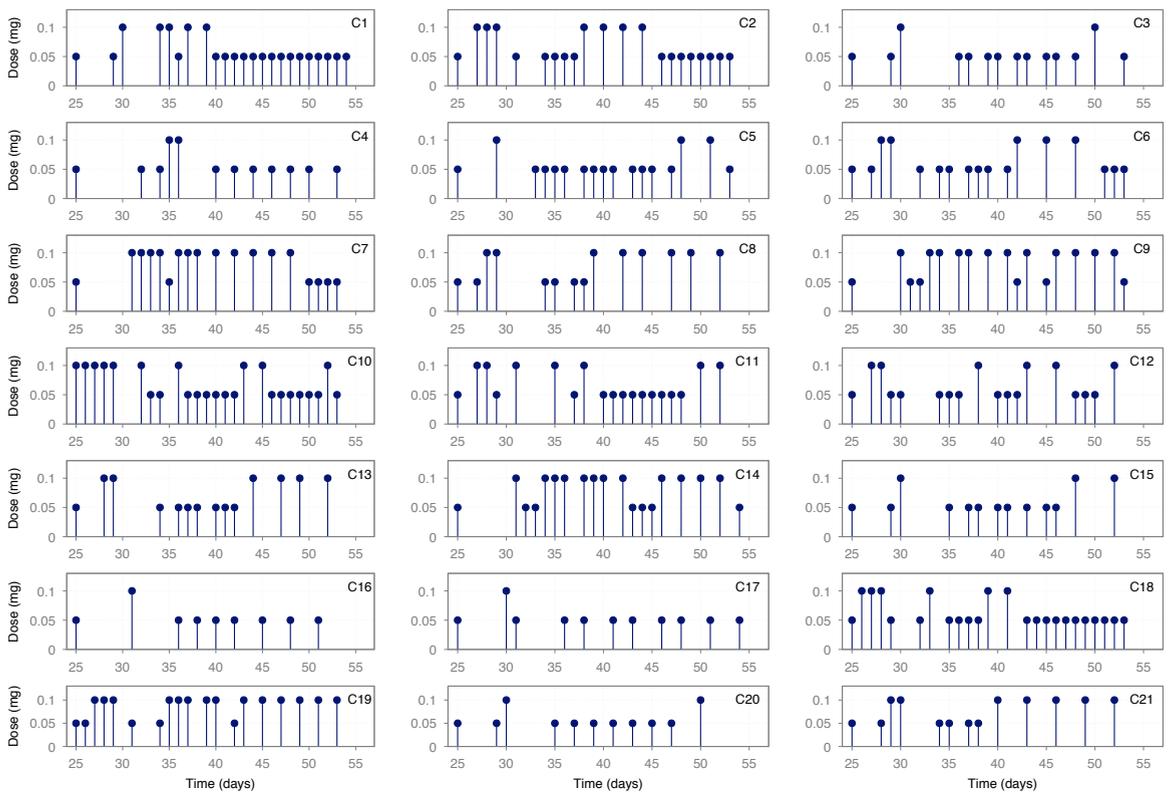

Figure 9: Personalised treatments computed through our multi-arm ISCT in 15 days.

Model-based approaches, exploiting PharmacoKinetics (PK), as *e.g.*, [55], are used instead to build populations of virtual phenotypes. Such populations are used to optimise and individualise drug doses [56, 57]. PK-based models, however, do not define how administered drugs can affect a Virtual Phenotype (VP) (namely, PharmacoDynamics, PD), *i.e.*, possible side-effects due to drug administrations are not taken into account.

In our model-based setting, we have to face with complex Virtual Physiological Human (VPH) models, *e.g.*, HumMod [19], Physiomodel [20], and GynCycle [6] defined through highly non-linear differential equations modelling the underlying biological mechanisms (*e.g.*, inhibitory and stimulatory effects). As outlined in Section 2.1, such VPH models are hybrid systems that can be defined by systems of Ordinary Differential Equations (ODEs) (see, *e.g.*, [58, 32, 33]) whose inputs are discrete event sequences (see, *e.g.*, [41, 44]). To find an optimal treatment means to find an optimal plan in hybrid domains, where the behaviour of the given system is described by both discrete and continuous quantities. In the literature, there are many techniques to model planning problems in hybrid domains, *e.g.*, PDDL+ [59, 60]. Most of PDDL+ planners can deal only with linear dynamics (*e.g.*, [61]). [62] proposed a Satisfiability Modulo Theories (SMT)-based approach for solving PDDL+ problems with non-linear dynamics. Model checking techniques are also used to find plans. Examples in this direction are [63, 64], which exploits symbolic model checking, UPMurphi [65, 66], which given as input a PDDL+ problem specification computes a universal plan, CGMurphi [67], an explicit model checker used to compute optimal controllers, and [68, 69, 70, 71], which define methodologies to compute controllers for non-linear hybrid systems. However, as outlined in Section 4.2, the typical complexity of the differential equations of VPH models relevant for clinical practice makes such models out of reach for symbolic approaches like those mentioned above, and appoints numerical integration as the only viable means to compute (black-box) the model evolutions under a given input function.

In particular, even considering that clinical actions have constant and equal duration (as, *e.g.*, in [72]), no reasoning or inference can be made on action effects in a black-box setting as ours, because the only way to interact with the models is through numerical simulation.

The automated synthesis of rational decisions and plans in black-box environments is common in several other application domains of high industrial relevance, like smart grids (see, *e.g.*, [73, 74, 75]), games (see, *e.g.*, [76]) and real-time manoeuvring of Unmanned Aerial Vehicles (see, *e.g.*, [77]).

The works closest to ours are those in [76, 78] and citations thereof, where a simulator is used to discover the effect of actions. In such works, the simulator is defined as a factored state model where actions are (black-box) procedures and states are represented in terms of variables. Then an algorithm, namely, Iterated Width (IW), is typically employed. It consists of iteratively calling a breadth-first procedure, namely, $\texttt{IW}(i)$, with $i = 1, 2, 3, \ldots$, until the problem is solved or $i$ becomes greater than the number of state variables (see, *e.g.*, [79]). During each call to $\texttt{IW}(i)$, states are pruned accordingly to how novel they are. A state is novel if and only if values for at most $i$ state variables have not seen before. Authors point out that IW is efficient for the majority of classical planning problems where it is enough that $i = 1, 2$, *i.e.*, actions change at most 2 state variables. However, in our setting, our simulator state consists of the union of all state vectors, *i.e.*, $\mathbf{x}(t)$, of each VP+monitor within the input digital twin. Moreover, the most of variables within such state vectors consist of real-valued biological quantities. Also, we remind that the dynamics of such model are defined by means of ODEs. Hence, when we apply a clinical action (and simulate our digital twin) the vast majority of state variables change their value. In this setting, applying `IW` means nearly always to call only `IW`($n$), with $n$ the overall number of variable of our simulator state. This behaviour makes the pruning on which `IW` relies on quite ineffective as `IW` degenerates to prune truly duplicate states, which are also extremely rare as our variables are real valued. Note that, in order to increase pruning of duplicate states, one could think of simplifying the dynamics of the VPH model at hand (*e.g.*, by discretising state variable domains). However, this is not a straightforward approach as it poses several questions about accuracy, credibility, and trust of the entailed model predictions, which have to be again experimentally assessed.

## 7. Conclusions

In this paper we presented methods and algorithms based on intelligent search aimed at synthesising optimal personalised treatments by means of In Silico Clinical Trials (ISCT), exploiting quantitative models of the physiology and drugs Pharmacokinetics/Pharmacodynamics (PKPD) of interest, and clinical measurements on human patients from which we define their digital twins.

We applied our approach on a case study involving a complex state-of-the-art model of the human female Hypothalamic–Pituitary–Gonadal (HPG) axis, in order to compute, for any given patient, a *personalised treatment* for the downregulation phase of an assisted reproduction protocol, which is effective on the patient at hand, but minimises the overall amount of drug used (hence, indirectly, the associated cost as well as likelihood and severity of adverse effects).

The possibility to *optimise in silico*, in a few weeks of computation on a High Performance Computing (HPC) infrastructure, a complex treatment for a *given* human patient *before* its actual administration shows the potential of artificial intelligence for model-based (*in silico*) precision medicine. This however calls for *trusted* Virtual Physiological Human (VPH) models, which is currently one of the major obstacles for the uptake of ISCT in clinical practice.

Indeed, the results of our ISCT (conducted using a state-of-the-art *validated* VPH model as GynCycle) are extremely promising, but must be taken with care. In particular, an *in vivo* evaluation of the actual effectiveness of the personalised treatments generated by our algorithm is needed in order to assess the accuracy of GynCycle in predicting the patient reactions to *non-standard drug dosing patterns*, as those computed by our algorithm.

### Acknowledgements

We are grateful to the co-authors of the preliminary work in [1].

This work was partially supported by the following research projects/grants: Sapienza University 2018 project RG11816436BD4F21 "Computing Complete Cohorts of Virtual Phenotypes for In Silico Clinical Trials and Model-Based Precision Medicine"; Italian Ministry of University & Research (MIUR) grant "Dipartimenti di Eccellenza 2018–2022" (Dept. Computer Science, Sapienza Univ. of Rome); EC FP7 project PAEON (Model Driven Computation of Treatments for Infertility Related Endocrinological Diseases, 600773); INdAM "GNCS Project 2019"; CINECA Class C ISCRA Project no. HP10CPJ9LW.